\documentclass{article}
\usepackage{arxiv}

\usepackage[utf8]{inputenc} 
\usepackage[T1]{fontenc}    
\usepackage{hyperref}       
\usepackage{url}            
\usepackage{booktabs}       
\usepackage{amsfonts}       
\usepackage{nicefrac}       
\usepackage{microtype}      
\usepackage{lipsum}		
\usepackage{graphicx}
\usepackage{natbib}
\usepackage{doi}

\usepackage{amsmath}
\usepackage{xspace}
\usepackage{xcolor,colortbl}
\usepackage{subcaption}

\title{Autoassociative Learning of Structural Representations for Modeling and Classification in Medical Imaging}
\author{\href{https://orcid.org/0009-0004-3922-1975}{\includegraphics[scale=0.06]{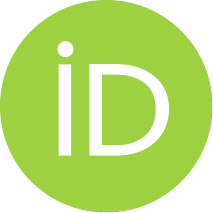}\hspace{1mm}}Zuzanna Buchnajzer \and \href{https://orcid.org/0009-0000-2309-1003}{\includegraphics[scale=0.06]{orcid.pdf}\hspace{1mm}}Kacper Dobek \and \href{https://orcid.org/0009-0002-8182-8485}{\includegraphics[scale=0.06]{orcid.pdf}\hspace{1mm}}Stanisław Hapke \and \href{https://orcid.org/0009-0007-3070-1632}{\includegraphics[scale=0.06]{orcid.pdf}\hspace{1mm}}Daniel Jankowski \and \href{https://orcid.org/0000-0001-5439-3231}{\includegraphics[scale=0.06]{orcid.pdf}\hspace{1mm}}Krzysztof Krawiec\thanks{\url{https://ml.cs.put.poznan.pl/en}} \\
Institute of Computing Science\\
Poznan University of Technology, Poznan, Poland\\
\texttt{krawiec@cs.put.poznan.pl} \\
}

\renewcommand{\undertitle}{Manuscript under consideration at the International Journal of Applied Mathematics and Computer Science}
\renewcommand{\shorttitle}{Autoassociative Learning of Structural Representations for Modeling and Classification}

\hypersetup{
pdftitle={Autoassociative Learning of Structural Representations for Modeling and Classification},
pdfsubject={q-bio.NC, q-bio.QM},
pdfauthor={Zuzanna Buchnajzer, Kacper Dobek, Stanisław Hapke, Daniel Jankowski, Krzysztof Krawiec},
pdfkeywords={First keyword, Second keyword, More},
}

\begin{document}

\newcommand{\mname}{ASR\xspace} %
\maketitle


\begin{abstract}
Deep learning architectures based on convolutional neural networks tend to rely on continuous, smooth features. While this characteristics provides significant robustness and proves useful in many real-world tasks, it is strikingly incompatible with the physical characteristic of the world, which, at the scale in which humans operate, comprises crisp objects, typically representing well-defined categories. This study proposes a class of neurosymbolic systems that learn by reconstructing images in terms of visual primitives and are thus forced to form high-level, structural explanations of them. When applied to the task of diagnosing abnormalities in histological imaging, the method proved superior to a conventional deep learning architecture in terms of classification accuracy, while being more transparent. 
\end{abstract}


\section{Introduction}

The blueprint of convolutional neural networks (convnets), the working horse of deep learning (DL) applications in computer vision (CV), assumes that raster-based processing at consecutive layers gives rise to higher-level features that convey information about objects, the structures they form, and other aspects of scene content. While this proved effective in many use cases, the structural reasoning in convnets is only \emph{implicit} in the sense of residing in weights in units' receptive fields. In particular, convnets have no means of explicitly capturing the `objectness' of percepts. This is strikingly at odds with the physical characteristic of the world, which, at the scale in which humans operate, is full of objects that have well-defined properties such as shape, size, orientation, and color. 

This incompatibility with the characteristics of natural images has several downsides. Firstly, it increases the risk of overfitting, as the combinatorial capacity of raster-based processing is excessively expressive for capturing natural scenes. To curb this risk, convnets need to be trained on huge volumes of data, which can be particularly troublesome in supervised settings that require image annotation. Last but not least, they offer little explanatory capability and need specialized algorithms to elucidate the processing or justify the decisions being made.  

To address the above challenges,  we propose \mname, a neurosymbolic autoencoder that forms Auto-associative Structural Representations, physically plausible scene descriptions that capture and explain the observed image 
in terms of \emph{visual primitives} rather than individual pixels. This representation is explicit, i.e.\ the model is forced to reason in the prescribed terms by autoassociative learning: given an input image, \mname attempts to reproduce it by parameterizing the primitives. The architecture comprises a convolutional encoder queried on a raster image and a symbolic decoder that performs differentiable rendering of primitives to reproduce that image. As all these components are differentiable, the architecture can be efficiently trained end-to-end with conventional gradient-based algorithms, using a simple pixel-wise reconstruction loss that compares the reproduced image with the input image. 

\mname can work with any visual primitives that can be rendered in a differentiable fashion. However, to provide alignment with the characteristics of natural images and to improve explainability, it is desirable for the primitives to be spatially continuous, compact, and tunable with parameters that independently control human-interpretable visual properties, such as size, orientation, shape, and color. In this study, we resort to ellipses as a simple yet nontrivial class of primitives. We demonstrate the approach by applying it to microscopic images of stained histological sections (Sec.\ \ref{sec:experiment}). The \mname is first trained on the (unannotated) images via autoassociation. Then, we use the learned representation to build a transparent diagnostic model (a decision tree), demonstrate its predictive superiority to a baseline configuration, and show how the decisions it makes can be explained with the interpretable properties of visual primitives. 


\begin{figure}
    \centering
    \includegraphics[width=\linewidth]{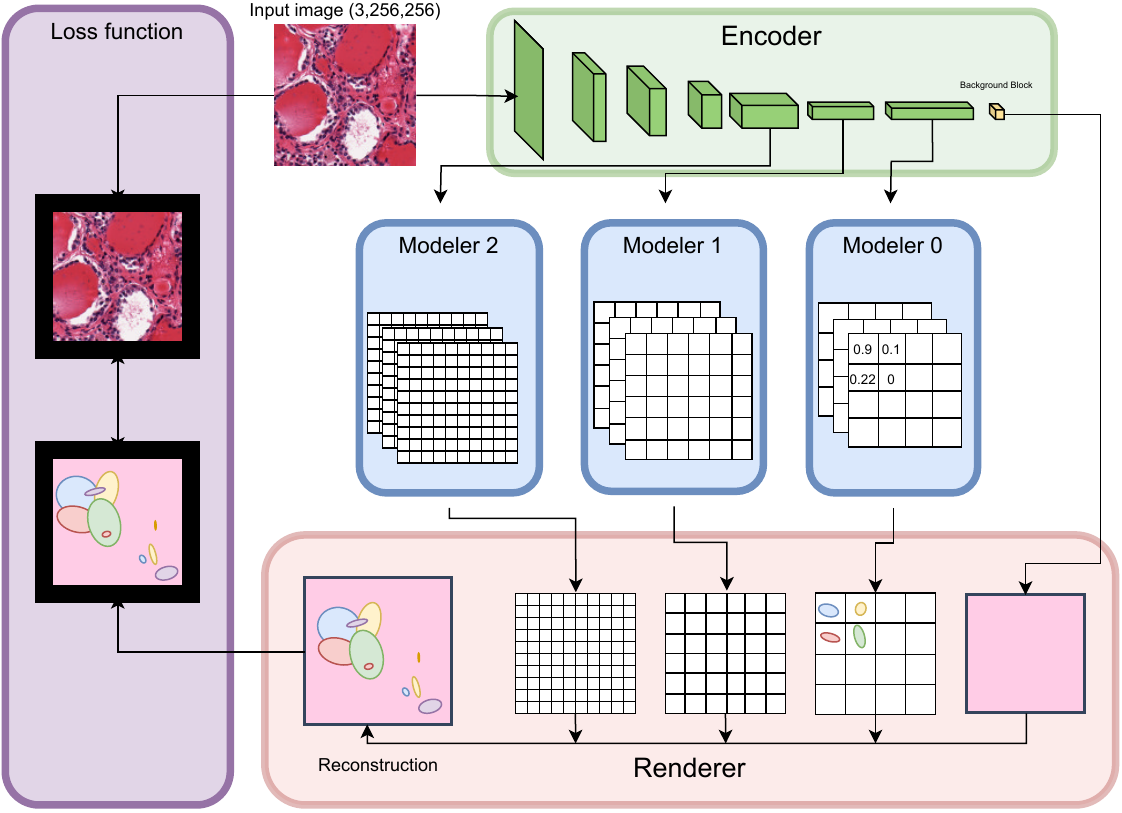}
    \caption{The overview of \mname's architecture.}
    \label{fig:architecture-overview}
    \vspace{-5mm}
\end{figure}


\section{ASR architecture and training}\label{sec:architecture}

The \mname autoencoder comprises an Encoder, a number of Modelers, and a Renderer -- the last one corresponding to the decoder in conventional autoencoders (Fig.\ \ref{fig:architecture-overview}).  

\vspace{1mm}\noindent\textbf{Encoder.}
The Encoder is a convnet implemented as a stack of \emph{ConvBlocks}, each comprising a 2D convolutional layer, ReLU activation function, and batch normalization. Consecutive ConvBlocks have decreasing spatial dimensions (spatial scales) and an increasing number of output channels. The stack of ConvBlocks is concluded with a \emph{BackgroundBlock}, which flattens the output of the last ConvBlock and maps it with a small dense network to three variables $r_{bg}$, $g_{bg}$, $b_{bg}$ that are interpreted as the color components of background in later processing stages. 

\vspace{1mm}\noindent\textbf{Modelers.}
In contrast to conventional autoencoders that process only the final output of the encoder, \mname makes use of multiple latents formed by ConvBlocks at several spatial scales. Let $z_j$ denote the latent vector produced by the $j$th ConvBlock at a given location in the input image (the spatial grid). The task of the Modeler is to map $z_j$ to human-interpretable parameters of the graphical primitive to be rendered by the Renderer. 

There is a separate modeler for each spatial scale $j$ (Fig.\ \ref{fig:architecture-overview}), realized as a trainable $1\times1$ convolution layer that maps $z_j$ at each location in the image to six variables that control the rendering of the primitive at that location: the horizontal and vertical scaling factors ($w_j$ and $h_j$), the rotation angle $d_j$, and the tuple of RGB components $a_j$ that control the primitive's color appearance. The layer features the sigmoid activation function to bound the output intervals. Subsequently, $w_j$ and $h_j$ are linearly scaled to $[0.1, 2]$, so that it becomes impossible to `collapse' the primitive to a point, while the upper bound of 2 allows partial overlap of primitives. $d_j$ is scaled to $[0, 2\pi]$. 

As argued later, the quality of reproduction that guides the training process of \mname is only a means to an end of \emph{learning informative structural representations}. Therefore, the absolute values of that metric are not very relevant. For this reason, and to avoid excessive computational cost, we resort to \emph{spatially sparse rendering} of visual primitives, rather than allowing them to be `anchored' at any pixel in the image. We achieve this with two means. Firstly, as in many convnets, the dimensions of the spatial latent decrease with scale, so that the receptive fields of individual elements of those latents, expressed in terms of the input image, become larger with consecutive scales. Secondly, each Modeler is applied to its spatial latent with a stride greater than 1. As a result, the primitives at scale $j$ are produced at certain spacing $r_j$ and overlap only when the scaling factors $w_j$ and $h_j$ are relatively high. Technically, we set $r_j$ to half the size of the receptive field at scale $j$. 


\vspace{1mm}\noindent\textbf{Renderer.}
The Renderer produces the output image based on the parameters received from the Modeler. This involves three stages: rendering of individual primitives at scale $j$, combining them into a single resultant canvas at scale $j$, and merging these canvases into the final reproduction. 

As \mname is meant to be trained end-to-end with gradient, the rendering process must be differentiable. While prior work on differentiable rendering exists (see, e.g., \citep{ravi2020pytorch3d,Li:2020:DVG,KaolinLibrary,Mitsuba3}), the elliptical shapes of our primitives allow achieving this goal with a relatively crude rendering of a blurry `blobs' rather than `crisp' ellipses. For a given ellipse at scale $j$, we realize this with the following steps: 
\begin{enumerate}
    \item Create a monochrome (single-channel) zeroed raster $R_j$ of size $(2r_j+1) \times (2r_j+1)$. 
    \item Draw a centered blurry circle with radius $r_j$ by setting the brightness of each pixel $p \in R_j$ to $\sigma(\beta(r_j-d(p)))$, where $d(p)$ is the distance of $p$ from the center of $R_j$,  $\sigma$ is the sigmoid function, and $\beta$ is the fuzziness coefficient that controls the sharpness of the contour.
    \item Construct the affine transformation matrix $A$ based on the scaling coefficients ($w_j$, $h_j$) and rotation $d_j$ predicted by the Modeler. 
    \item Apply $A$ to $R_j$: generate a transformed grid of points from $A$ and use it to sample from $R$ with bilinear interpolation (so this step is differentiable too).  
    \item Multiply $R_j$ by the appearance descriptor $a_j$ to produce an RGB image of the same dimensions. 
\end{enumerate}
Notice that the resulting $R$ is approximately double the size of the grid cell, which allows the rendered primitives to overlap if necessary, as signaled earlier.

The $R_j$s for all grid cells are subsequently fused into a single canvas that represents the rendering outcome at the $j$th scale. To this aim, the individual $R$s are spatially co-registered, i.e.\ placed at grid cells in the coordinate system of the canvas, and their pixel values are aggregated. As in this study, we are interested in imaging in transmissive mode, where the light is \emph{absorbed} by objects (rather than reflected or emitted by them), we perform multiplicative aggregation of the complements of $R$s in each RGB channel independently, so that the resultant channel value of a pixel $(x,y)$ in the canvas is $\Pi_i (1-R_j^{(i)}(x,y))$
where $R_j^{(i)}(x,y)$ is the channel value of pixel $(x,y)$ in $i$th raster at scale $j$, after spatial co-registration.

The final reproduction is obtained by element-wise multiplication of the canvases obtained for particular scales $j$, followed by the multiplication with the background canvas, which is uniformly filled with the RGB values predicted by the BackgroundBlock. 

Notice that the Renderer does not involve any trainable parameters.



\vspace{1mm}\noindent\textbf{Training.}
\mname is trained with gradient descent by minimizing the loss function defined as the mean square error between the reproduction produced by the Renderer and the input image, averaged over pixels and RGB channels. We ignore the margin of 16 pixels in this calculation, which translates to approximately 6\% of the image width,  to lessen the impact of border effects (the model may be struggling to reproduce the partially visible structures that extend beyond the image). We refer to this cropped variant of MSE as Masked MSE (MMSE). 

Preliminary experiments indicated the importance of parameter initialization, which is due to the highly nonlinear processing taking place in the Renderer and the presence of sigmoid functions in Modelers. Therefore, we initialize the weights of ConvBlocks and Modelers with the Xavier method \citep{pmlr-v9-glorot10a}, but sample their biases from $N(0,1)$ and set the bias of the BackgroundBlock to 1 to encourage it to participate in the rendering process.

\section{Related work}

\mname represents the category of image understanding systems inspired by the "\emph{vision as inverse graphics}" blueprint \citep{barrow1978recovering}, which can be seen as a computer vision instance of the broader \emph{analysis-by-synthesis} paradigm. While considered in the CV community for decades (see, e.g., \citep{Krawiec_2007}), it experienced significant advancement in recent years, thanks to the rapid progress of DL that facilitated end-to-end learning of complex, multi-staged architectures. Below, we review selected representatives of this research thread; for a thorough review of other approaches to compositional scene representation via reconstruction, see \citep{Yuan_Chen_Li_Xue_2023}; also, \cite{Elich_2022} present a compact review of numerous works on related CV topics, including learning object geometry and multi-object scene representations, segmentation, and shape estimation. 

The Multi-Object Network (MONet) proposed by \cite{Burgess_2019} is a composite unsupervised architecture that combines image segmentation based on an attention mechanism (to delineate image components) with a variational autoencoder (VAE), for rendering individual components in the scene. As such, the approach does not involve geometric aspects of image formation and scene understanding. Also, it does not involve geometric rendering of objects: the subimages of individual components are generated with the VAE and `inpainted' into the scene using raster masks.  

PriSMONet by \cite{Elich_2022} attempts decomposition of 3D scenes based on their 2D views. Similarly to MONet, it parses the scene sequentially, object by object, and learns to generate objects' views composed from several aspects: shape, textural appearance, and 3D extrinsics (position, orientation, and scale). The background is handled separately. Object shapes are represented using the Signed Distance Function formalism, well known in CV, and generated using a separate autoencoder submodel (DeepSDF); in this sense, PriSMONet does not involve shape priors. In contrast to \citep{Burgess_2019}, the architecture engages differentiable rendering. 

Another related research direction concerns part discovery, where the goal is to decompose the objects comprising the scene into constituents, which preferably should have well-defined semantics (i.e., segmentation at the part level, not the object level). The approaches proposed therein usually rely on mask-based representations (see, e.g. \citep{Hung2019,Choudhury2022}); some of them involve also geometric transforms (e.g. \citep{Hung2019}). 

\cite{10.1007/978-3-031-71167-1_13} recently proposed a related neurosymbolic architecture that uses a domain-specific language to capture selected priors of image formation, including object shape (represented by elliptic Fourier transform), appearance, categorization, and geometric transforms. They express template programs in that language and learn their parameterization with features extracted from the scene by a convnet. When executed, the parameterized program produces geometric primitives which are rendered in a differentiable fashion.  


\section{Experimental validation}\label{sec:experiment}

The experimental validation presented in the following comprises representation learning (Stage 1, Sec.\ \ref{sec:stage1}) and a `downstream' task of learning to classify (Stage 2, Sec.\ \ref{sec:stage2}). In the former, we train \mname and baseline models (all autoencoders) on a training set of images via autoassociation, to allow them to form informative latent representations. In Stage 2, we compose the Encoders and Modelers extracted from the models obtained in Stage 1 with decision trees to perform image classification. The main research question is whether the structure-aware representations formed by \mname in Stage 1 translate in Stage 2 to better test-set accuracy than the baseline models.

\subsection{Problem statement and data}\label{sec:data}

\begin{figure}[t!]
    \begin{subfigure}{0.32\columnwidth}
        \caption{Benign} 
        \includegraphics[width=\linewidth]{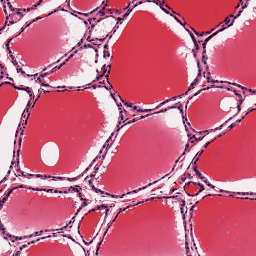}
        \label{fig: Benign sample}
    \end{subfigure}
    \hfill
    \begin{subfigure}{0.32\columnwidth}
        \caption{Hashimoto}
        \includegraphics[width=\linewidth]{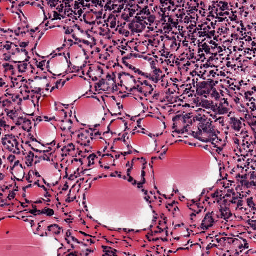}
        \label{fig: Hashimoto sample}
    \end{subfigure}
    \hfill
    \begin{subfigure}{0.32\columnwidth}
        \caption{Nodularity}
        \includegraphics[width=\linewidth]{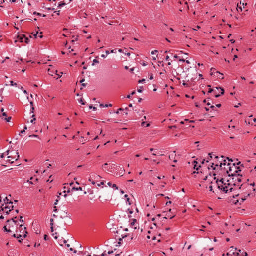}
        \label{fig: nodularity sample}
    \end{subfigure}
    \vspace{-5mm}
    \caption{Examples of patches extracted from three different WSIs of the thyroid gland. In the Benign class (a) the follicles are regularly distributed across the tissue and have a circular shape. Aside from the vesicles, not much connective tissue is visible. The Hashimoto's disease (b) features lymphoplasmacytic infiltration, which manifests as numerous dark purple-stained cells (lymphocytes) in the interfollicular area, while the follicles are smaller and sparsely distributed. In Nodularity (c), we observe a small number of thyroid follicles and a high proportion of connective tissue.}
    \label{figs: patches descriptions}
    \vspace{-5mm}
\end{figure}

As the challenges outlined in the Introduction (data inefficiency, costly annotation, limited explainability) are particularly relevant for medical imaging, we focus on this realm. Given that \mname, as described in Sec.\ \ref{sec:architecture}, is designed to reconstruct images specifically with elliptical shapes, we choose microscopic histological imaging, where cells and other structures can be expected to lend themselves to modeling with such primitives. More specifically, we consider images of the human thyroid as the subject of this study. Thyroid follicles filled with colloid hormones form the major mass of the organ, and are usually oval, or almost round; see examples in Figs.\  \ref{figs: patches descriptions} and \ref{fig:gtex-sample}. Yet they vary in shapes, sizes, orientations, colors, and spatial arrangements, which makes them a perfect yardstick for the kind of modeling implemented by \mname. 

We rely on the Biospecimen Research Database \citep{BiospecimenResearchDatabase} as the source of examinations for this study (Sec.\ \ref{sec:data}). From this collection, we selected a sample of 30 examinations representing three clinical conditions (decision classes): \emph{Benign} (no abnormalities), \emph{Hashimoto}, and \emph{Nodularity} (Fig.\ \ref{figs: patches descriptions}). When sampling, we provided balanced distributions of sex and age of patients (Table \ref{tab:dataset-urls}). The class assignment is based on the golden standard, as present in the clinical review comments included in the metadata.  

We divide the 30 selected examinations into the training set (15 examinations), the validation set (6 examinations) and the test set (9 examinations). This partitioning was stratified, so that these proportions hold also in each decision class. Each examination comes from a different patient. 


WSIs are too large to process them directly with DL models and, as exemplified in Fig.\ \ref{fig:gtex-sample}, usually contain a substantial fraction of white, uninformative background. Therefore, we extract patches from WSIs by scanning them systematically with a 1024$\times$1024 sliding window at horizontal and vertical stride of 635 pixels (2/3 of window size) and accept only the patches that are occupied with tissue in at least 80\%. Then, we downsize them to $256\times256$ pixels. This results in from 400 to 1550 256$\times$256 patches per WSI. As a result, the train:validation:test proportions of 15:6:9 on the level of examinations translate to $10{,}915$:$4{,}945$:$7{,}235$ (i.e. approximately 15:6.8:9.9) on the level of image patches. As a side-effect, the distributions of decision classes in these subsets are not uniform and diverge significantly (Fig.\ \ref{fig:balance}).  

\subsection{Configuration of \mname}

We configure \mname so that its encoder yields latent from three ConvBlocks (Sec.\ \ref{sec:architecture}) and modeling and reconstruction take place at scales/resolutions corresponding to those blocks: the highest (finest, $j=2$, grid spacing $r_2=128$), the intermediate ($j=1$, $r_1=64$), and the lowest (coarsest, $j=0$, $r_0=32$). The grid dimensions at those scales are, respectively, 2$\times$2 cells, 4$\times $4 cells, and 8$\times$8 cells, so that the encoder parameterizes 84 ellipses in total. As each of them requires 6 parameters, the overall dimensionality of the structural latent amounts to 504. 
All models are designed to process the downsized $256\times 256$ patches and trained until the PyTorch's EarlyStopping component signaled stagnation on the validation set (with patience of 20 epochs), for the number of epochs listed in \ref{sec:config} alongside other settings.

\subsection{Compared variants of the method}

We compare three variants of \mname and one baseline configuration, detailed in the following. 

\vspace{1mm}\noindent\textbf{Base}. The base variant of the method, as presented in Sec.\ \ref{sec:architecture}, trained with the MMSE reconstruction loss for up to 50 epochs. 

\vspace{1mm}\noindent\textbf{Regularized}. In preliminary experimenting, \mname exhibited a significant tendency to rely primarily on rendering at higher resolutions (on the denser grids), while neglecting the lowest resolutions. As this is at odds with our goal of learning succinct and transparent representations, we devise the \emph{Appearance Regularization Value} to curb the number of graphical primitives used in rendering:
\begin{equation}
\label{appearance_loss}
ARV(\hat{y})=\frac{1}{|Loc_J|}\sum_{j \in J} w_j \sum_{l \in Loc_j} \sum_{c \in {a_{j}^l}} c^\alpha, 
\end{equation}
where
$\hat{y}$ is the reconstruction image,
$J$ is the set of all scales,
$Loc_J$ is the set of variables describing shapes for all scales in $\hat{y}$,
$Loc_j$ is the set of variables describing shapes at scale $j$,
$w_j$ is the weight of scale $j$,
$c$ is the value corresponding to RGB color descriptors at scale $j$ and location $l$, and
$\alpha$ is the factorization parameter $\in (0.5, 1.0)$. Based on preliminary experimenting, we set the scale weights as follows: $w_0=0.6$, $w_1=0.9$, and $w_2=1.2$.
The trade-off between this regularization term and the quality of reconstruction is controlled by weighing ARV in the total loss:
\begin{equation}
    \label{reg-loss-formula}
    Loss(y, \hat{y}) = \textrm{MMSE}(y, \hat{y}) + \lambda_a ARV(\hat{y}),
\end{equation}
where we set $\lambda_a$ to 0.009 in the course of preliminary tuning. 
Similarly to the Base configuration, Regularized models are trained for 50 epochs.

\vspace{1mm}\noindent\textbf{Incremental}. While the Regularized models indeed turned out to render at high resolution more sparingly than the Base configurations, we sought other means of addressing their over-representation using incremental training, in which the model is forced to rely more on the coarser resolutions earlier in training, while with time being gradually allowed to use the higher-resolution primitives. This is achieved by multiplying all outputs of the Modeler for scale $j$ by $\beta_j$. We start with $\beta_0=1$ and $\beta_1=\beta_2=0.01$. Then, $\beta_j$s are increased after each epoch according to the formula
$\mathit{\beta_j \leftarrow \min(1.0, \beta_j + \gamma_j).}$
Based on preliminary experimenting, we set $\gamma_1=\gamma_2=0.1$ and $\gamma_0=0$, i.e.\ the rendering at the lowest resolution is unaffected by this mechanism and remains constant throughout training. The incrementing commences  in epoch 8 for scale 1 and in epoch 20 for scale 2. 

The Incremental configuration engages also regularization described in the Regularized variant, albeit only from the 35th epoch, to avoid the potentially unstable interference of both mechanisms. Training lasts for up to 55 epochs, i.e.\ 5 epochs more than the previous variants, to give the regularization mechanism enough time to influence the model. 

\vspace{1mm}\noindent\textbf{Baseline}. As a baseline, we use a conventional convolutional autoencoder with a global latent. Its encoder starts with a stack of encoder blocks, each comprising a convolutional layer, an Exponential Linear Unit (ELU) activation function, batch normalization, and a 2D max pooling layer. This is followed by flattening to a vector and a single dense layer that produces the global latent. We set the dimensionality of the global latent to 200 as a compromise between the computational cost and reconstruction quality. The decoder is a stack of decoder blocks, each comprising a 2D up-sampling layer with nearest neighbor interpolation, a convolutional layer with ELU activation function, and batch normalization. As for \mname, the input and output are 256$\times$256 RGB images. The architecture has $3{,}781{,}509$ trainable parameters and has been trained for up to 50 epochs. 

\subsection{Stage 1: Image reconstruction}\label{sec:stage1}

In this stage, we train all configurations on the image patches derived from the 15 training examinations, to perform image reconstruction, i.e. the diagnostic labels are ignored. For each configuration, we conduct 5 runs that start with a different initialization of the model's weights. 

Table \ref{table:results-table} compares the performance of configurations in terms of four reconstruction metrics comparing the produced output image to the input image, all calculated on the test set: Mean Absolute Error (MAE), Mean Square Error (MSE), Structural Similarity Index Measure (SSIM) by \cite{Wang2004-lr}, and Masked MSE (MMSE), i.e. the actual reconstruction loss used in training (Sec.\ref{sec:architecture}). 

\begin{table}[t!]
\centering
\caption{The performance of individual runs of three variants of \mname and the Baseline autoencoder. The model with the best MSE within a configuration is marked in bold.}
\footnotesize
\begin{tabular}{llllc}
\hline
\textbf{Model} & \textbf{MSE \textdownarrow} & \textbf{MAE \textdownarrow} & \textbf{SSIM \textuparrow} & \textbf{MMSE \textdownarrow} \\ \hline
Base\_1 & 0.0342 & 0.1393 & 0.1936 & 0.0342 \\
Base\_2 & 0.0330 & 0.1370 & 0.1943 & 0.0330 \\
Base\_3 & 0.0327 & 0.1362 & 0.1938 & 0.0326 \\
Base\_4 & 0.0329 & 0.1367 & 0.1924 & 0.0329 \\
\textbf{Base\_5} & \textbf{0.0326} & \textbf{0.1363} & \textbf{0.1921} & \textbf{0.0325} \\
\hline
\textit{Mean} & 0.0331 & 0.1371 & 0.1932 & 0.0330 \\ \hline
Regularized\_1 & 0.0341 & 0.1415 & 0.1968 & 0.0341 \\
\textbf{Regularized\_2} & \textbf{0.0341} & \textbf{0.1417} & \textbf{0.1976} & \textbf{0.0341} \\
Regularized\_3 & 0.0350 & 0.1431 & 0.1972 & 0.0350 \\
Regularized\_4 & 0.0342 & 0.1407 & 0.1976 & 0.0342 \\
Regularized\_5 & 0.0345 & 0.1412 & 0.1961 & 0.0345 \\
\hline
\textit{Mean} & 0.0344 & 0.1416 & 0.1971 & 0.0344 \\ \hline
\textbf{Incremental\_1} & \textbf{0.0342} & \textbf{0.1415} & \textbf{0.1956} & \textbf{0.0342} \\
Incremental\_2 & 0.0347 & 0.1424 & 0.1958 & 0.0347 \\
Incremental\_3 & 0.0349 & 0.1428 & 0.1968 & 0.0349 \\
Incremental\_4 & 0.0366 & 0.1466 & 0.1961 & 0.0366 \\
Incremental\_5 & 0.0346 & 0.1420 & 0.1960 & 0.0346 \\
\hline
\textit{Mean} & 0.0350 & 0.1431 & 0.1961 & 0.0350 \\ \hline
Baseline\_1 & 0.0274 & 0.1178 & 0.2299 & - \\
Baseline\_2 & 0.0276 & 0.1187 & 0.2271 & - \\
Baseline\_3 & 0.0272 & 0.1179 & 0.2295 & - \\
\textbf{Baseline\_4} & \textbf{0.0271} & \textbf{0.1174} & \textbf{0.2292} & \textbf{-} \\
Baseline\_5 & 0.0278 & 0.1202 & 0.2227 & - \\
\hline
\textit{Mean} & 0.0274 & 0.1184 & 0.2277 & - \\ \hline
\end{tabular}
\label{table:results-table}
\end{table}

\begin{figure*}[t!]
    \centering
    \includegraphics[width=0.8\linewidth]{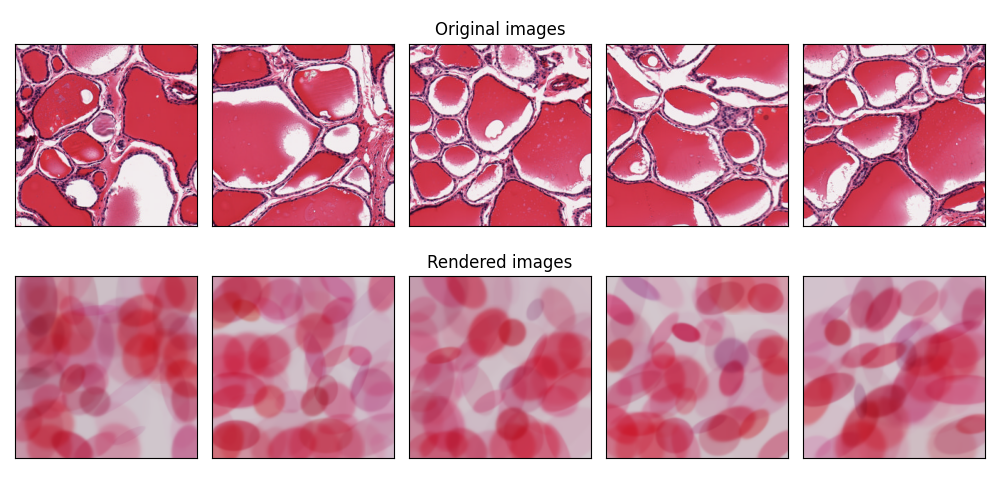}\vspace{-5mm}
    \caption{Reconstructions of sample test images produced by the Regularized\_2 model.}
    \label{fig:rec-reg-2}
    \vspace{-4mm}
\end{figure*}

The values of MMSE are systematically very close to those of MSE, which suggests that \mname copes well with border effects. Nevertheless, \mname configurations fare worse than the Baseline on all metrics. However, the differences are not large; for instance, on MAE, the best \mname configurations lag behind the mean Baseline by less than 0.02 (the maximum per-pixel MAE is 1). Among the \mname configurations, it is interesting to observe that both the Regularized and Incremental variants improve upon Base on SSIM while deteriorating on MAE and MSE. This suggests that they tend to pay more attention to the structures present in images, rather than individual pixels, which aligns with the motivations of this study. These observations are to some extent corroborated with the visual representation of reconstructions, shown in Fig.\ \ref{fig:rec-reg-2}. 

\subsection{Stage 2: Classification}\label{sec:stage2}

In this stage, we use the encoders of models trained in Stage 1 to obtain diagnostic models that classify the image patches into the three considered classes (Benign, Hashimoto, Nodularity). 

\vspace{1mm}\noindent\textbf{Data preparation}. 
To increase the number of examples available for training and evaluation of classifiers (compared to the bare 30 examinations), we conduct this experiment on \emph{bags} of image patches. Each bag is a set of 16 patches drawn randomly (without replacement) from the same examination and inherits its decision class label. This results in $1{,}463$ bags, with train:validation:test proportions of 694:312:457. Figure \ref{fig: tabular_distribution} 
presents the distribution of decision classes in those subsets which, unsurprisingly, follows the distribution shown in the rightmost inset in Fig.\ \ref{fig:balance}. 

The encoding parts of compared architectures produce image features that are subsequently used to train and test a classifier. For the three \textbf{\mname} variants, this process comprises two steps for a given bag $B$:
    (i) Querying the Encoder and the Modelers on each of the sixteen 256$\times$256 image patches from $B$, each resulting with vectors of latent features per scale and location in grids.  
    (ii) Aggregation of the latent features simultaneously over space (the X and Y dimensions of the spatial latent) and over $B$, \emph{for each scale (0, 1, 2) and each latent variable independently}.  
    As there are 6 ellipse-controlling variables at each scale, there are 18 variables in total. These are then aggregated independently using mean and standard deviation, resulting in 36 features.   
The result of the above process is a tabular dataset of 1{,}463 bags (examples, rows) described by 36 attributes (columns) and partitioned into the three considered decision classes. Division into training, validation and test subsets remains  the same. 

For the \textbf{Baseline} autoencoder, we proceed analogously: the encoder is queried on all image patches from $B$, and the resulting latent dimensions are aggregated over $B$. However, given the high dimensionality of this latent and its global character (it captures the entire $256\times 256$ patch), we use only arithmetic mean for aggregation, applying it to each latent dimension independently. Therefore, for the Baseline, the resulting tabular dataset has dimensions 1{,}463 bags $\times$ 200 attributes (columns). 


\vspace{1mm}\noindent\textbf{The induction algorithm}. 
To demonstrate the explanatory potential of \mname, we choose decision trees as the predictive model to be trained on the tabular datasets prepared in the way described above (DecisionTreeClassifier implementation from the scikit-learn library \citep{scikit-learn}, an optimized version of the CART algorithm by \cite{Breiman1984ClassificationAR}). For fairness of comparison, all induced classifiers are subject to extensive hyperparameter tuning, detailed in \ref{sec:statistics}.

\vspace{1mm}\noindent\textbf{Results}. 
We produce a decision tree for each \mname and Baseline model listed in Table \ref{table:results-table}, and asses its predictive capacity on the testing part of the tabular dataset. Table \ref{table: classification-table} presents the accuracy of classification, precision, recall and F1 score of each of those classifiers. Notably, all configurations of \mname outperform the Baseline models on all metrics, and do so by a large margin. The best Baseline model is not better than the worst \mname model on any metric. One may argue that this may be due the Baseline models having over 5 times more attributes at their disposal (200 vs. 36). However, the extensive hyperparameter tuning (\ref{sec:statistics}) provided fair footing for the Baseline configurations, allowing them to pick the optimal hyperparameter values and pruning intensity. Apparently, however, this was not sufficient for Baseline to perform well: the `anonymous' features of its latent representation were not informative enough to provide for good predictive capacity. \mname, in contrast, managed to provide the decision trees with useful diagnostic information -- even though at the time of Stage 1 training (Sec.\ \ref{sec:stage1}) it was oblivious to the existence of decision classes. 

\begin{table}[t!]
\centering
\caption{Test-set predictive performance of each model. The model with the best accuracy and F1-score within a configuration is marked in bold.}
\footnotesize
\begin{tabular}{@{}llllc@{}}
\hline
\textbf{Model} & \textbf{Accuracy \textuparrow} & \textbf{F1\_Score \textuparrow} & \textbf{Precision \textuparrow} & \textbf{Recall \textuparrow} \\ \hline
Base\_1        & 0.5864   & 0.5048                   & 0.6245                   & 0.5864                \\
\textbf{Base\_2}        & \textbf{0.7768}   & \textbf{0.7735}                   & 0.8096                   & 0.7768                \\
Base\_3        & 0.7112   & 0.7038                   & 0.7164                   & 0.7112                \\
Base\_4        & 0.5930   & 0.5523                   & 0.5907                   & 0.5930                \\
Base\_5        & 0.7615   & 0.7414                   & 0.7957                   & 0.7615                \\ \hline
Regularized\_1         & 0.6980   & 0.6738                   & 0.7044                   & 0.6980                \\
Regularized\_2         & 0.7133   & 0.7040                   & 0.7040                   & 0.7133                \\
Regularized\_3         & 0.6827   & 0.6655                   & 0.6762                   & 0.6827                \\
Regularized\_4         & 0.6674   & 0.6785                   & 0.7079                   & 0.6674                \\
\textbf{Regularized\_5}         & \textbf{0.7133}   & \textbf{0.7075}                   & 0.7245                   & 0.7133                \\ \hline
Incremental\_1 & 0.7090   & 0.6892                   & 0.7471                   & 0.7090                \\
Incremental\_2 & 0.6958   & 0.6998                   & 0.7058                   & 0.6958                \\
Incremental\_3 & 0.6521   & 0.6131                   & 0.7061                   & 0.6521                \\
\textbf{Incremental\_4} & \textbf{0.7505}   & \textbf{0.7511}                   & 0.7568                   & 0.7505                \\
Incremental\_5 & 0.7396   & 0.7339                   & 0.7651                   & 0.7396                \\ \hline
Baseline\_1     & 0.4486   & 0.3572                   & 0.2979                   & 0.4486                \\
Baseline\_2     & 0.4311   & 0.3797                   & 0.3945                   & 0.4311                \\
Baseline\_3     & 0.2801   & 0.2342                   & 0.3729                   & 0.2801                \\
Baseline\_4     & 0.3129   & 0.2776                   & 0.3928                   & 0.3129                \\
\textbf{Baseline\_5}     & \textbf{0.5383}   & \textbf{0.4486}                   & 0.4108                   & 0.5383                \\ \hline
\end{tabular}
\vspace{-5mm}
\label{table: classification-table}
\end{table}

Statistical analysis of Table \ref{table: classification-table} (ANOVA and post-hoc Holm test) indicates  significant superiority of each variant of \mname compared to the Baseline, on accuracy of classification and on the F1 score (see Sec.\ \ref{sec:statistics} details). There are no significant differences between the variants of \mname. Notably, the Base variant exhibits most variance between runs, while the Regularized one is most stable in this respect. 

\begin{figure*}[t!]
    \centering
    \includegraphics[width=\linewidth]{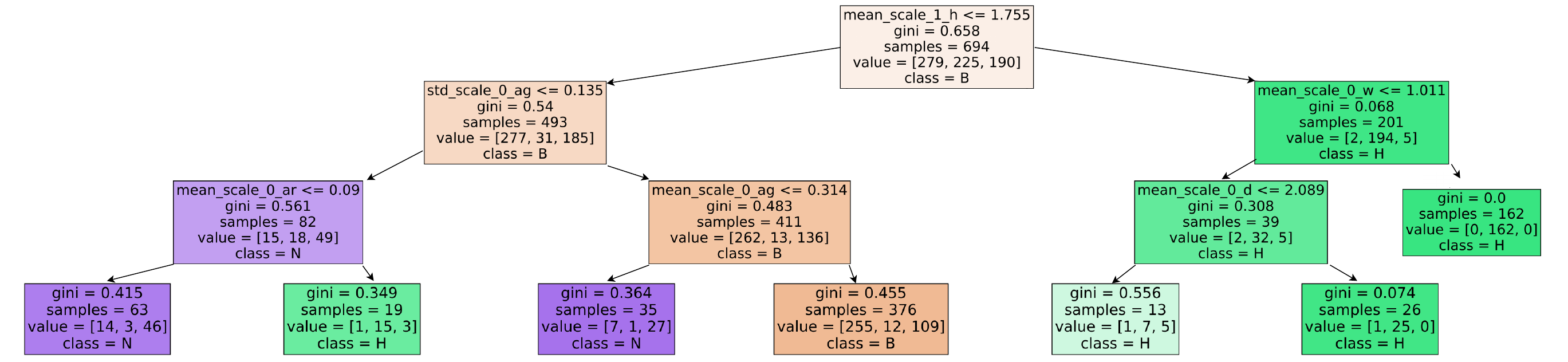}
    \caption{The final pruned decision tree induced from the attributes provided by the Base\_2 model.}
\label{fig: best_tree}
\vspace{-4mm}
\end{figure*}

Figure \ref{fig: best_tree} presents the decision tree that achieved the best accuracy of classification across all \mname runs (Base\_2, the second run of the Base configuration). Similarly to other trees induced from the \mname-generated latent representation (not shown here for brevity), this tree nicely delineates most examples of the Hashimoto decision class (H) already at the root of the tree, using the mean of the height ($h$) ellipse parameter at scale $j=1$. The remaining two decision classes (N, Normal and B, Benign) turn out to be much harder to differentiate. Overall, the Nodularity class turns out to be hardest to classify, which we attribute to the relatively large fraction of connective tissue and sparse, small blobs, which are difficult to reconstruct using only ellipses (Fig.\ \ref{figs: patches descriptions}). Additionally, the examinations from this decision class occasionally feature inflammation, which can be mismatched with some visual features of Hashimoto. 

More importantly, the obtained decision tree is notably small, comprising only 6 decision nodes and 7 leaves\footnote{It is easy to notice that the pruning algorithm failed to fold the right-hand Hashimoto subtree, which would further reduce these numbers}. Trees induced from attributes provided by other ASR instances were similarly compact. This corroborates the main hypothesis of this study, i.e. that the structural, compositional representation enforced by \mname correlates better with relevant visual features, and provides a more reliable basis for classification.

Interestingly, out of the five unique attributes used by decision nodes, only one (the mean height used by the root node) comes from scale $j=1$, and none come from the scale $j=2$. This signals that the features of large, coarse ellipses built at the lowest scale are most useful for this particular diagnostic scenario. To verify this conjecture, we performed feature importance analysis by assessing the gain of the Gini criterion for each attribute (a.k.a. Gini importance), and averaged it over the final, optimized and pruned decision trees obtained from all 15 instances of \mname variants from Table \ref{table:results-table}). Figure \ref{fig: features-importance_heatmap} shows the result of this analysis, which confirms that scale 0 features are most important, followed by scale 1 and then scale 2. In the group of color features ($a_r$, $a_g$, $a_b$), the standard deviation of the green color component at scale 0 turns out to be particularly informative; this may be caused by the transmissive nature of the imaging process, in which changes in the green component result in different shades of purple. Among the geometric features, the standard deviation of the orientation angle of the ellipse (\emph{d}) at scale 0 proves most informative. 

The transparency of \mname can be further exploited by tracing the decision-making process back to the specific instances of ellipses in the input image, i.e.\ those that supported the specific condition(s) on the path traversed in the decision tree. 


\section{Conclusions}

We provided preliminary evidence for the usefulness of a new representation learning approach, in which the learner is forced to `explain' the data in terms of higher-level concepts. This allows forming compact, informative and explainable representations, which are not only capable of conveying the essence of the content of the input image, but also constructing potent predictors with high explanatory power. 

\mname offers natural means for providing a learning system with domain knowledge in the form of the repertoire of visual primitives. For the class of images considered in this study, elliptical shapes proved sufficient to outperform a conventional DL baseline. It is likely, however, that more sophisticated graphical representations could further improve the quality of reconstruction and the accuracy of the resulting classifiers; for instance, the Fourier-based shape representations used in \citep{10.1007/978-3-031-71167-1_13} offer more flexibility in shape modeling, and thus could likely convey more relevant information. 

\newpage\appendix
\section{The data}\label{sec:data}

\begin{figure}[t!]
    \centering
    \includegraphics[width=\columnwidth]{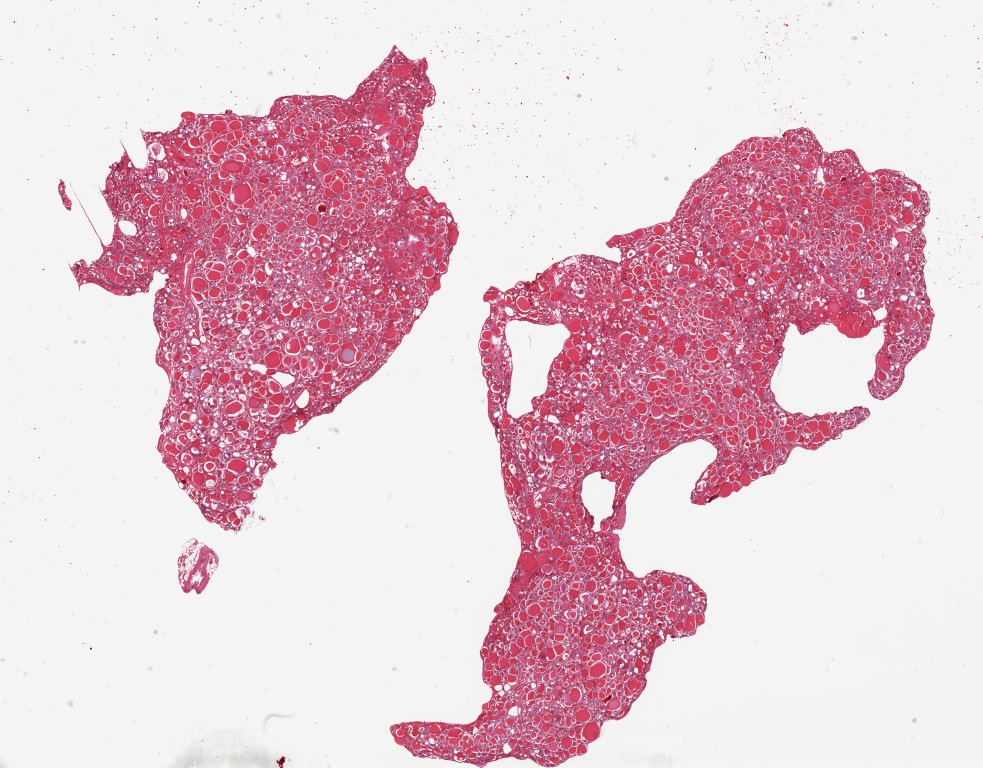}
    \caption{Example Thyroid Gland Whole Slide Image from the Biospecimen Research Database (BRD) \cite{BiospecimenResearchDatabase}.}
    \label{fig:gtex-sample}
\end{figure}

The Biospecimen Research Database  \citep{BiospecimenResearchDatabase} (BRD) that served as the source of data in this study is a large repository of digital slides containing, among others, 893 Whole Slide Images (WSIs) of the thyroid gland, each at least 50k $\times$ 50k pixels in size and typically covering a specimen of at least 10 by 10 mm (see example slide in Fig.\ \ref{fig:gtex-sample}). The slides are available in Aperio SVS format, which we parsed using the \emph{slideio} library \citep{choosehappy_2020} that allows reading in both image data and metadata. Though dimensions of images vary significantly in the sample, their spatial resolution is the same (0.4942 $\mu$m/pix). The apparent magnification (AppMag) is also constant across the dataset (20). Figure \ref{fig:gtex-sample} presents an example of Thyroid Gland WSI from the BRD. 

Table \ref{tab:dataset-urls} lists the examinations selected for the dataset used in the experimental part of this study. The selected sample is publicly available under the following address: \url{https://tinyurl.com/asr-gtex}. 

Figure \ref{fig: tabular_distribution} presents the distribution of patients' attributes in the selected sample, including factorization into decision classes.

\begin{table*}[h!]
    \caption{The 30 digital slides from the BRD database \cite{BiospecimenResearchDatabase} of GTEx images used in this study. These examinations were selected using the search keyword ``Thyroid gland``. Each entry contains the image in SVS format and metadata, including the case ID, sex, and age (presented below). Another field in the database contains pathology review comments, for instance, ``2 pieces, no abnormalities``. This global information was used to determine the class of the sample (seen in the Class column). In total, ten images per class were selected -- five for training, two for validation, and three for testing. The last column describes the number of patches extracted from a digital slide.}
    \resizebox{\textwidth}{!}{%
    \begin{tabular}{@{}
    >{\columncolor[HTML]{FFFFFF}}l 
    >{\columncolor[HTML]{FFFFFF}}l 
    >{\columncolor[HTML]{FFFFFF}}l 
    >{\columncolor[HTML]{FFFFFF}}l 
    >{\columncolor[HTML]{FFFFFF}}l 
    >{\columncolor[HTML]{FFFFFF}}l 
    >{\columncolor[HTML]{FFFFFF}}r @{}}
    \toprule
    {\color[HTML]{333333} \textbf{Case ID}} &
      {\color[HTML]{333333} \textbf{Class}} &
      {\color[HTML]{333333} \textbf{Sex}} &
      {\color[HTML]{333333} \textbf{Age}} &
      {\color[HTML]{333333} \textbf{URL}} &
      {\color[HTML]{333333} \textbf{Subset}} &
      \multicolumn{1}{l}{\cellcolor[HTML]{FFFFFF}{\color[HTML]{333333} \textbf{Patches}}} \\ \midrule
    {\color[HTML]{333333} GTEX-1128S} &
      {\color[HTML]{333333} Benign} &
      {\color[HTML]{333333} Female} &
      {\color[HTML]{333333} 61-70} &
      {\color[HTML]{333333} {\url {https://brd.nci.nih.gov/brd/specimen/GTEX-1128S-0126}}} &
      {\color[HTML]{333333} train} &
      {\color[HTML]{333333} 1171} \\
    {\color[HTML]{333333} GTEX-1192X} &
      {\color[HTML]{333333} Benign} &
      {\color[HTML]{333333} Male} &
      {\color[HTML]{333333} 51-60} &
      {\color[HTML]{333333} {\url {https://brd.nci.nih.gov/brd/specimen/GTEX-1192X-1126}}} &
      {\color[HTML]{333333} train} &
      {\color[HTML]{333333} 1246} \\
    {\color[HTML]{333333} GTEX-12KS4} &
      {\color[HTML]{333333} Benign} &
      {\color[HTML]{333333} Male} &
      {\color[HTML]{333333} 21-40} &
      {\color[HTML]{333333} {\url {https://brd.nci.nih.gov/brd/specimen/GTEX-12KS4-0526}}} &
      {\color[HTML]{333333} train} &
      {\color[HTML]{333333} 500} \\
    {\color[HTML]{333333} GTEX-13NYB} &
      {\color[HTML]{333333} Benign} &
      {\color[HTML]{333333} Male} &
      {\color[HTML]{333333} 41-50} &
      {\color[HTML]{333333} {\url {https://brd.nci.nih.gov/brd/specimen/GTEX-13NYB-0726}}} &
      {\color[HTML]{333333} train} &
      {\color[HTML]{333333} 774} \\
    {\color[HTML]{333333} GTEX-13PL6} &
      {\color[HTML]{333333} Benign} &
      {\color[HTML]{333333} Female} &
      {\color[HTML]{333333} 41-50} &
      {\color[HTML]{333333} {\url {https://brd.nci.nih.gov/brd/specimen/GTEX-13PL6-1026}}} &
      {\color[HTML]{333333} train} &
      {\color[HTML]{333333} 679} \\
    {\color[HTML]{333333} GTEX-13QBU} &
      {\color[HTML]{333333} Benign} &
      {\color[HTML]{333333} Female} &
      {\color[HTML]{333333} 21-40} &
      {\color[HTML]{333333} {\url {https://brd.nci.nih.gov/brd/specimen/GTEX-13QBU-0626}}} &
      {\color[HTML]{333333} val} &
      {\color[HTML]{333333} 993} \\
    {\color[HTML]{333333} GTEX-1A8G6} &
      {\color[HTML]{333333} Benign} &
      {\color[HTML]{333333} Male} &
      {\color[HTML]{333333} 61-70} &
      {\color[HTML]{333333} {\url {https://brd.nci.nih.gov/brd/specimen/GTEX-1A8G6-0626}}} &
      {\color[HTML]{333333} val} &
      {\color[HTML]{333333} 1252} \\
    {\color[HTML]{333333} GTEX-1AMFI} &
      {\color[HTML]{333333} Benign} &
      {\color[HTML]{333333} Female} &
      {\color[HTML]{333333} 51-60} &
      {\color[HTML]{333333} {\url {https://brd.nci.nih.gov/brd/specimen/GTEX-1AMFI-0526}}} &
      {\color[HTML]{333333} test} &
      {\color[HTML]{333333} 177} \\
    {\color[HTML]{333333} GTEX-14DAR} &
      {\color[HTML]{333333} Benign} &
      {\color[HTML]{333333} Male} &
      {\color[HTML]{333333} 41-50} &
      {\color[HTML]{333333} {\url {https://brd.nci.nih.gov/brd/specimen/GTEX-14DAR-0126}}} &
      {\color[HTML]{333333} test} &
      {\color[HTML]{333333} 715} \\
    {\color[HTML]{333333} GTEX-1F5PL} &
      {\color[HTML]{333333} Benign} &
      {\color[HTML]{333333} Female} &
      {\color[HTML]{333333} 41-50} &
      {\color[HTML]{333333} {\url {https://brd.nci.nih.gov/brd/specimen/GTEX-1F5PL-0826}}} &
      {\color[HTML]{333333} test} &
      {\color[HTML]{333333} 580} \\
    {\color[HTML]{333333} \textbf{Total}} &
      {\color[HTML]{333333} } &
      {\color[HTML]{333333} } &
      {\color[HTML]{333333} } &
      {\color[HTML]{333333} } &
      \multicolumn{1}{l}{\cellcolor[HTML]{FFFFFF}{\color[HTML]{333333} }} &
      {\color[HTML]{333333} \textbf{8087}} \\
    {\color[HTML]{333333} GTEX-14PJ6} &
      {\color[HTML]{333333} Hashimoto} &
      {\color[HTML]{333333} Female} &
      {\color[HTML]{333333} 61-70} &
      {\color[HTML]{333333} {\url {https://brd.nci.nih.gov/brd/specimen/GTEX-14PJ6-0326}}} &
      {\color[HTML]{333333} train} &
      {\color[HTML]{333333} 493} \\
    {\color[HTML]{333333} GTEX-14PJM} &
      {\color[HTML]{333333} Hashimoto} &
      {\color[HTML]{333333} Female} &
      {\color[HTML]{333333} 51-60} &
      {\color[HTML]{333333} {\url {https://brd.nci.nih.gov/brd/specimen/GTEX-14PJM-1326}}} &
      {\color[HTML]{333333} train} &
      {\color[HTML]{333333} 694} \\
    {\color[HTML]{333333} GTEX-XBEC} &
      {\color[HTML]{333333} Hashimoto} &
      {\color[HTML]{333333} Male} &
      {\color[HTML]{333333} 51-60} &
      {\color[HTML]{333333} {\url {https://brd.nci.nih.gov/brd/specimen/GTEX-XBEC-0926}}} &
      {\color[HTML]{333333} train} &
      {\color[HTML]{333333} 745} \\
    {\color[HTML]{333333} GTEX-1AX8Y} &
      {\color[HTML]{333333} Hashimoto} &
      {\color[HTML]{333333} Male} &
      {\color[HTML]{333333} 41-50} &
      {\color[HTML]{333333} {\url {https://brd.nci.nih.gov/brd/specimen/GTEX-1AX8Y-0126}}} &
      {\color[HTML]{333333} train} &
      {\color[HTML]{333333} 757} \\
    {\color[HTML]{333333} GTEX-14BMU} &
      {\color[HTML]{333333} Hashimoto} &
      {\color[HTML]{333333} Female} &
      {\color[HTML]{333333} 21-40} &
      {\color[HTML]{333333} {\url {https://brd.nci.nih.gov/brd/specimen/GTEX-14BMU-0226}}} &
      {\color[HTML]{333333} train} &
      {\color[HTML]{333333} 845} \\
    {\color[HTML]{333333} GTEX-1C6WA} &
      {\color[HTML]{333333} Hashimoto} &
      {\color[HTML]{333333} Male} &
      {\color[HTML]{333333} 61-70} &
      {\color[HTML]{333333} {\url {https://brd.nci.nih.gov/brd/specimen/GTEX-1C6WA-0626}}} &
      {\color[HTML]{333333} test} &
      {\color[HTML]{333333} 868} \\
    {\color[HTML]{333333} GTEX-11XUK} &
      {\color[HTML]{333333} Hashimoto} &
      {\color[HTML]{333333} Female} &
      {\color[HTML]{333333} 41-50} &
      {\color[HTML]{333333} {\url {https://brd.nci.nih.gov/brd/specimen/GTEX-11XUK-0226}}} &
      {\color[HTML]{333333} test} &
      {\color[HTML]{333333} 729} \\
    {\color[HTML]{333333} GTEX-YJ8A} &
      {\color[HTML]{333333} Hashimoto} &
      {\color[HTML]{333333} Male} &
      {\color[HTML]{333333} 21-40} &
      {\color[HTML]{333333} {\url {https://brd.nci.nih.gov/brd/specimen/GTEX-YJ8A-0226}}} &
      {\color[HTML]{333333} test} &
      {\color[HTML]{333333} 1 549} \\
    {\color[HTML]{333333} GTEX-13QJC} &
      {\color[HTML]{333333} Hashimoto} &
      {\color[HTML]{333333} Female} &
      {\color[HTML]{333333} 61-70} &
      {\color[HTML]{333333} {\url {https://brd.nci.nih.gov/brd/specimen/GTEX-13QJC-0826}}} &
      {\color[HTML]{333333} val} &
      {\color[HTML]{333333} 565} \\
    {\color[HTML]{333333} GTEX-WVJS} &
      {\color[HTML]{333333} Hashimoto} &
      {\color[HTML]{333333} Male} &
      {\color[HTML]{333333} 51-60} &
      {\color[HTML]{333333} {\url {https://brd.nci.nih.gov/brd/specimen/GTEX-WVJS-0726}}} &
      {\color[HTML]{333333} val} &
      {\color[HTML]{333333} 1 055} \\
    {\color[HTML]{333333} \textbf{Total}} &
      {\color[HTML]{333333} } &
      {\color[HTML]{333333} } &
      {\color[HTML]{333333} } &
      {\color[HTML]{333333} } &
      \multicolumn{1}{l}{\cellcolor[HTML]{FFFFFF}{\color[HTML]{333333} }} &
      {\color[HTML]{333333} \textbf{8300}} \\
    {\color[HTML]{333333} GTEX-1MCYP} &
      {\color[HTML]{333333} Nodularity} &
      {\color[HTML]{333333} Female} &
      {\color[HTML]{333333} 21-40} &
      {\color[HTML]{333333} {\url {https://brd.nci.nih.gov/brd/specimen/GTEX-1MCYP-0626}}} &
      {\color[HTML]{333333} train} &
      {\color[HTML]{333333} 444} \\
    {\color[HTML]{333333} GTEX-1KD5A} &
      {\color[HTML]{333333} Nodularity} &
      {\color[HTML]{333333} Male} &
      {\color[HTML]{333333} 51-60} &
      {\color[HTML]{333333} {\url {https://brd.nci.nih.gov/brd/specimen/GTEX-1KD5A-0426}}} &
      {\color[HTML]{333333} train} &
      {\color[HTML]{333333} 900} \\
    {\color[HTML]{333333} GTEX-15DCD} &
      {\color[HTML]{333333} Nodularity} &
      {\color[HTML]{333333} Female} &
      {\color[HTML]{333333} 61-70} &
      {\color[HTML]{333333} {\url {https://brd.nci.nih.gov/brd/specimen/GTEX-15DCD-1126}}} &
      {\color[HTML]{333333} train} &
      {\color[HTML]{333333} 574} \\
    {\color[HTML]{333333} GTEX-1I1GQ} &
      {\color[HTML]{333333} Nodularity} &
      {\color[HTML]{333333} Male} &
      {\color[HTML]{333333} 51-60} &
      {\color[HTML]{333333} {\url {https://brd.nci.nih.gov/brd/specimen/GTEX-1I1GQ-0826}}} &
      {\color[HTML]{333333} train} &
      {\color[HTML]{333333} 624} \\
    {\color[HTML]{333333} GTEX-131XW} &
      {\color[HTML]{333333} Nodularity} &
      {\color[HTML]{333333} Female} &
      {\color[HTML]{333333} 51-60} &
      {\color[HTML]{333333} {\url {https://brd.nci.nih.gov/brd/specimen/GTEX-131XW-0826}}} &
      {\color[HTML]{333333} train} &
      {\color[HTML]{333333} 469} \\
    {\color[HTML]{333333} GTEX-18QFQ} &
      {\color[HTML]{333333} Nodularity} &
      {\color[HTML]{333333} Male} &
      {\color[HTML]{333333} 21-40} &
      {\color[HTML]{333333} {\url {https://brd.nci.nih.gov/brd/specimen/GTEX-18QFQ-0726}}} &
      {\color[HTML]{333333} val} &
      {\color[HTML]{333333} 429} \\
    {\color[HTML]{333333} GTEX-13SLX} &
      {\color[HTML]{333333} Nodularity} &
      {\color[HTML]{333333} Female} &
      {\color[HTML]{333333} 51-60} &
      {\color[HTML]{333333} {\url {https://brd.nci.nih.gov/brd/specimen/GTEX-13SLX-0726}}} &
      {\color[HTML]{333333} val} &
      {\color[HTML]{333333} 651} \\
    {\color[HTML]{333333} GTEX-11EQ8} &
      {\color[HTML]{333333} Nodularity} &
      {\color[HTML]{333333} Male} &
      {\color[HTML]{333333} 61-70} &
      {\color[HTML]{333333} {\url {https://brd.nci.nih.gov/brd/specimen/GTEX-11EQ8-0826}}} &
      {\color[HTML]{333333} test} &
      {\color[HTML]{333333} 976} \\
    {\color[HTML]{333333} GTEX-130VJ} &
      {\color[HTML]{333333} Nodularity} &
      {\color[HTML]{333333} Female} &
      {\color[HTML]{333333} 51-60} &
      {\color[HTML]{333333} {\url {https://brd.nci.nih.gov/brd/specimen/GTEX-13OVJ-0626}}} &
      {\color[HTML]{333333} test} &
      {\color[HTML]{333333} 709} \\
    {\color[HTML]{333333} GTEX-11TUW} &
      {\color[HTML]{333333} Nodularity} &
      {\color[HTML]{333333} Male} &
      {\color[HTML]{333333} 61-70} &
      {\color[HTML]{333333} {\url {https://brd.nci.nih.gov/brd/specimen/GTEX-11TUW-0226}}} &
      {\color[HTML]{333333} test} &
      {\color[HTML]{333333} 932} \\
    {\color[HTML]{333333} \textbf{Total}} &
      {\color[HTML]{333333} } &
      {\color[HTML]{333333} } &
      {\color[HTML]{333333} } &
      {\color[HTML]{333333} } &
      \multicolumn{1}{l}{\cellcolor[HTML]{FFFFFF}{\color[HTML]{333333} }} &
      {\color[HTML]{333333} \textbf{6708}} \\ \bottomrule
    \end{tabular}%
    }
    \label{tab:dataset-urls}
    \end{table*}

\section{Configurations and hyperparameter settings} \label{sec:config}

The following parameters shared the same settings across all architectures considered in this study:
\begin{description}
\item[Optimizer:] The Adam optimizer \cite{kingma2017adam} has been used as it become the default choice in many deep learning frameworks due to its solid generalization performance across a wide range of tasks.
\item[Learning Rate = 0.001:] A starting point in many deep learning tasks. Across the experiments, several other values were tested, however, the initially used learning rate equal to 0.001 turned out to work the best for our case.
\item[Batch Size = 32:] A default value that is often chosen for memory efficiency, especially when training on GPUs.
\item[Number of batches per epoch = 32:] In most deep learning approaches, an \emph{epoch} refers to one complete cycle through the training data. However, in our implementation, it was decided to control the number of batches, and hence images processed by the learning algorithm in each iteration. Such a defined mechanism allowed the evaluation of the model performance more often. As a result, in one epoch $32\times32=1024$ images were passed through the model (the total size of the training set is 10,915 images).
\item[Number of epochs = 50 (55 for some configurations):] The maximum training duration for the model to show the convergence. In the configuration with the freezing of layers, it was decided to increase the number of epochs to 55. 
\item[Loss Function:] The Masked Mean Squared Error (MMSE) function described in Section \ref{sec:architecture} was used to not force the model to generate shapes that are only partially present in the input image.
\item[Early Stopping:] A default PyTorch \emph{EarlyStopping} handler with \emph{patience} = 20 was used to terminate the training if no improvement concerning the validation loss was seen after 20 epochs. It allowed us to reduce the computational costs and speed up the experimental process.
\item[Weight Initialization:] Xavier weight initialization \cite{pmlr-v9-glorot10a} has been used as it is a widely recommended choice for many neural network architectures, and more importantly, to prevent the vanishing gradient problem. See Table \ref{tab:weight-init} for initialization details per component.
\item[Seed:] To make different configurations more comparable, a fixed seed was set in PyTorch. It ensured that between particular training configurations, the order of the training data passed through the model, as well as the initial weights of the model, were the same.
\item[Data Augmentation:] While the built-in PyTorch module for image augmentation was initially incorporated as an optional training feature, offering support for vertical or horizontal flipping and rotating images at 90/180/270 degrees, it has been deactivated in all configurations due to a sufficient amount of data.
\end{description}

\begin{figure*}[t!]
    \begin{subfigure}{0.3\textwidth}
        \includegraphics[width=1\linewidth, trim={0 0 1.5cm 0}, clip]{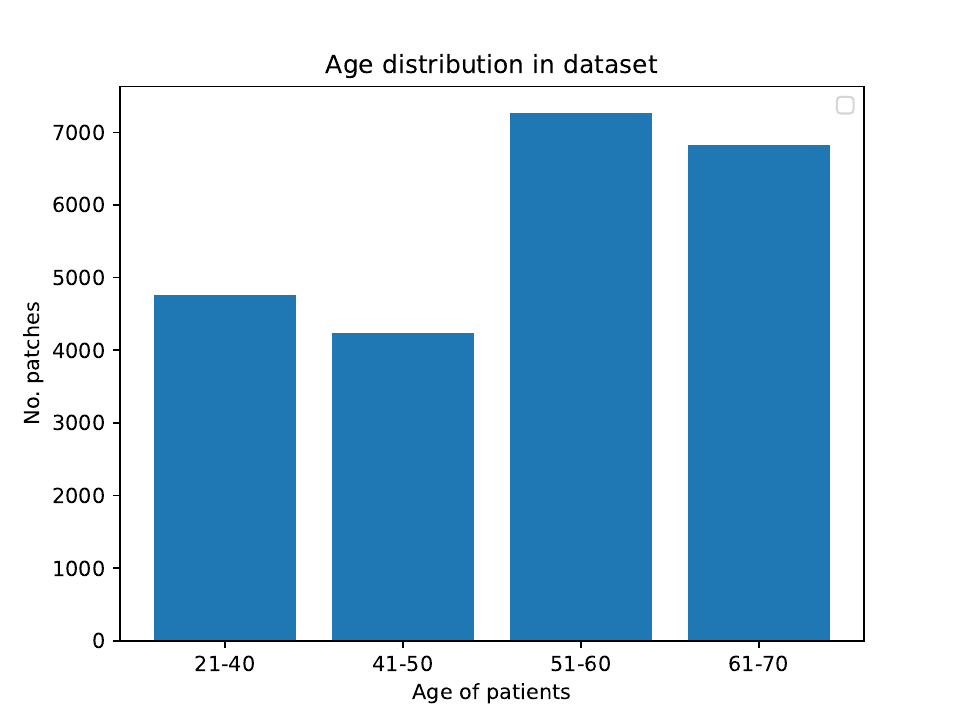}
        \label{fig: age balance}
    \end{subfigure}
    \hfill
    \begin{subfigure}{0.3\textwidth}
        \includegraphics[width=1\linewidth, trim={0 0 1.5cm 0}, clip]{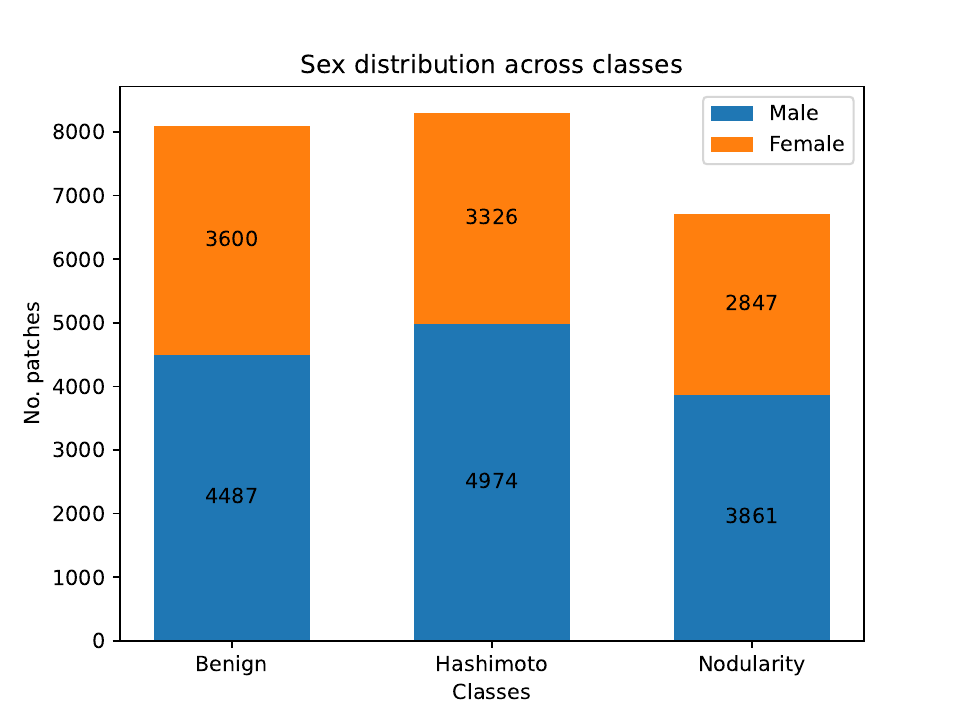}
        \label{fig :sex balance}
    \end{subfigure}
    \hfill
    \begin{subfigure}{0.3\textwidth}
        \includegraphics[width=1\linewidth, trim={0 0 1.5cm 0}, clip]{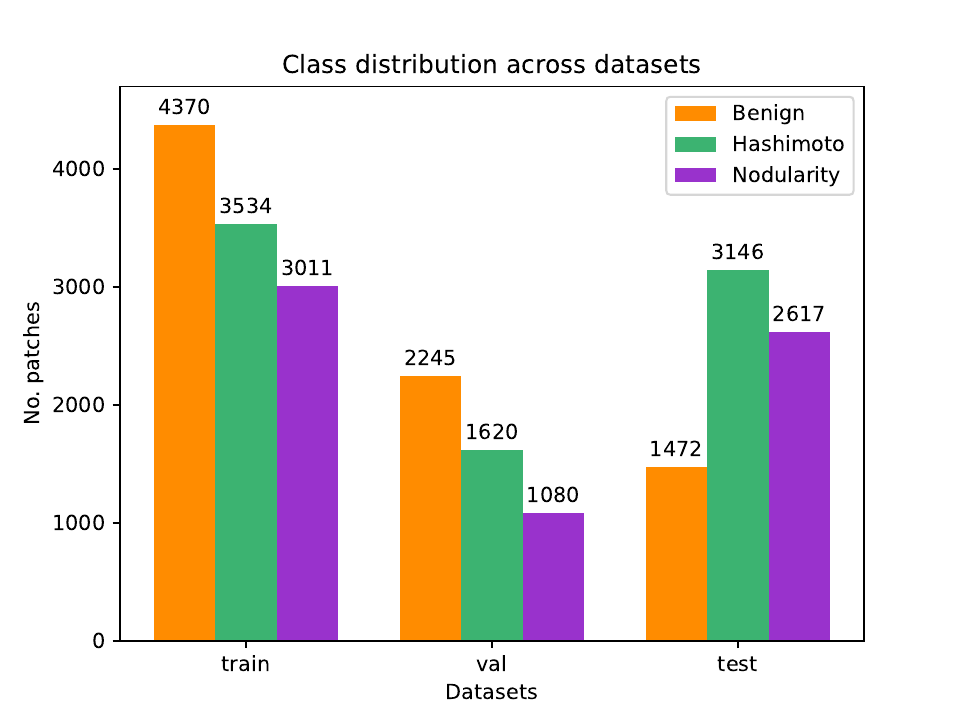}
        \label{fig: datasets_balance}
    \end{subfigure}
    \caption{The distribution of age (left) and sex and decision class (center) in the generated set of image patches. The right inset presents the distribution of decision classes after dataset partitioning into training, validation, and test sets. }
    \label{fig:balance}
\end{figure*}

\begin{figure}[t!]
    \centering
        \includegraphics[width=0.45\textwidth]{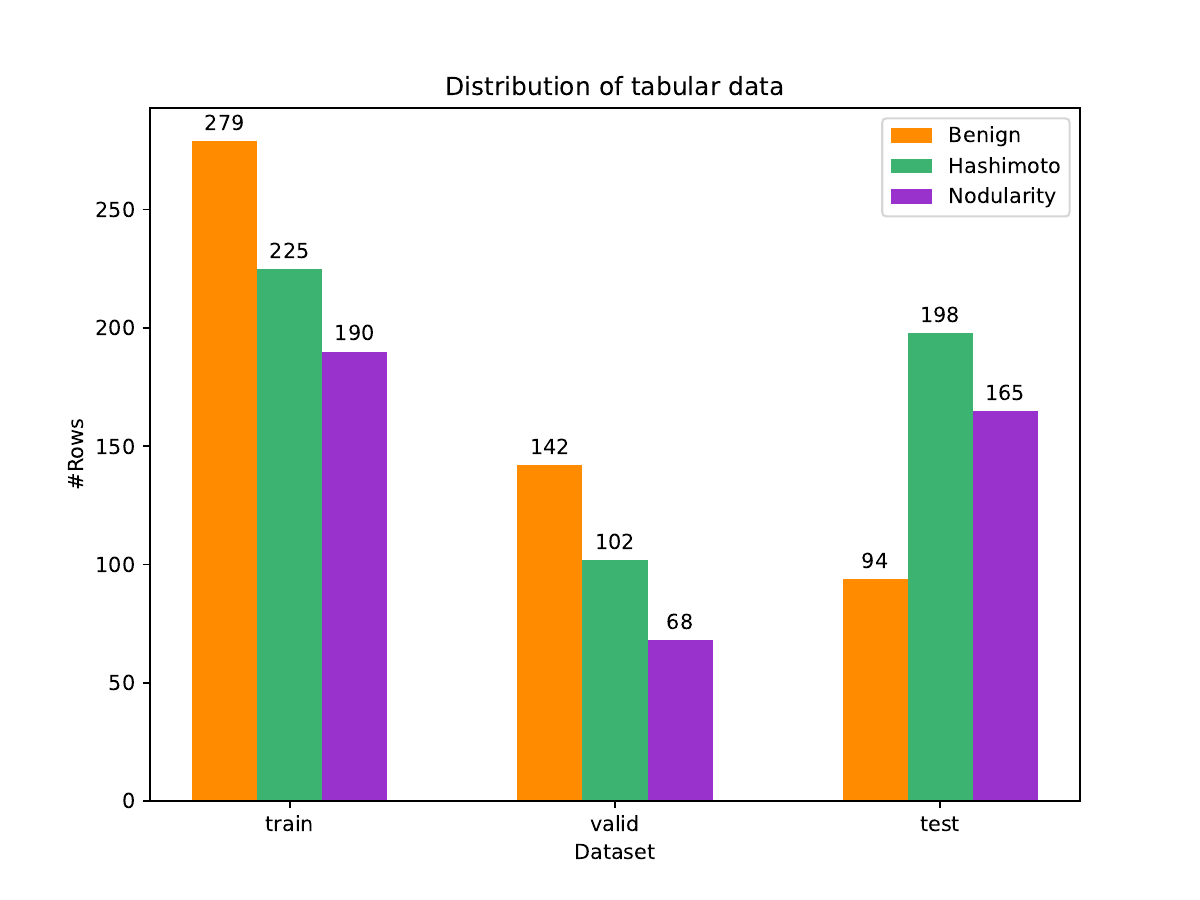}
    \caption{The distribution of classes in the tabular data is strongly correlated with the original distribution of the number of patches as shown in Figure \ref{fig:balance}.}
    \label{fig: tabular_distribution}
\end{figure}

In principle, the hyperparameters of \mname can be set arbitrarily. However, care should be taken when setting the grid spacing $r_j$. For the Modeler and the Renderer to be able to faithfully reproduce the input observed in a given part of the input image, the grid spacing at a given scale $j$ should be similar to the size of the receptive field of units in the corresponding ConvBlock. If the receptive field is smaller than grid spacing, the Renderer will attempt to produce objects based on only partial information available in its visual field, which may destabilize the training process. Conversely, if the receptive field is too big relative to grid spacing, the Renderer can be `frustrated' by being unable to render some large objects it observes.

\begin{table}[t!]
\centering
\caption{Weight initialization method by component and layer.}
\begin{tabular}{@{}llll@{}}
\toprule
Component       & Layer  & Weights & Bias   \\ \midrule
ConvBlock       & Conv2d & Xavier  & Normal \\
Modeler         & Conv2d & Xavier  & Normal \\
BackgroundBlock & Linear & Xavier  & Ones   \\ \bottomrule
\end{tabular}
\label{tab:weight-init}
\end{table}

\section{Classification (Stage 2)}\label{sec:statistics}

We optimize the tree induction algorithm by exhaustively iterating over 30 combinations of hyperparameter values: node impurity measure (\emph{gini}, \emph{entropy}), maximum height of the tree (3, 4, 5, 7, $\infty$), and the minimum number of samples per leaf (5, 10, 20). For each of those combinations, the accuracy of classification is assessed using a 5-fold cross validation conducted on the training part of the tabular dataset. The combination that maximizes this measure is subsequently used to build a decision tree from the entire training set; however, to further maximize the predictive capability, this tree undergoes adaptive pruning using the minimal cost-complexity method. The intensity of pruning is controlled by the ccp\_alpha parameter of DecisionTreeClassifier, with 0 being a neutral value (no pruning) and larger values causing folding of terminal subtrees into leaves. We tune ccp\_alpha to maximize the classification accuracy on the validation set. The tree pruned in this way is subsequently tested on the test set.  

Figure \ref{fig: tabular_distribution} presents the distribution in the tabular dataset created in Stage 2 of the experiment, to demonstrate its high similarity to the distribution of image patches presented in Fig.\ \ref{fig:balance}.

\begin{figure*}[t!]
    \centering
    \begin{subfigure}{0.6\textwidth}
        \includegraphics[width=\textwidth]{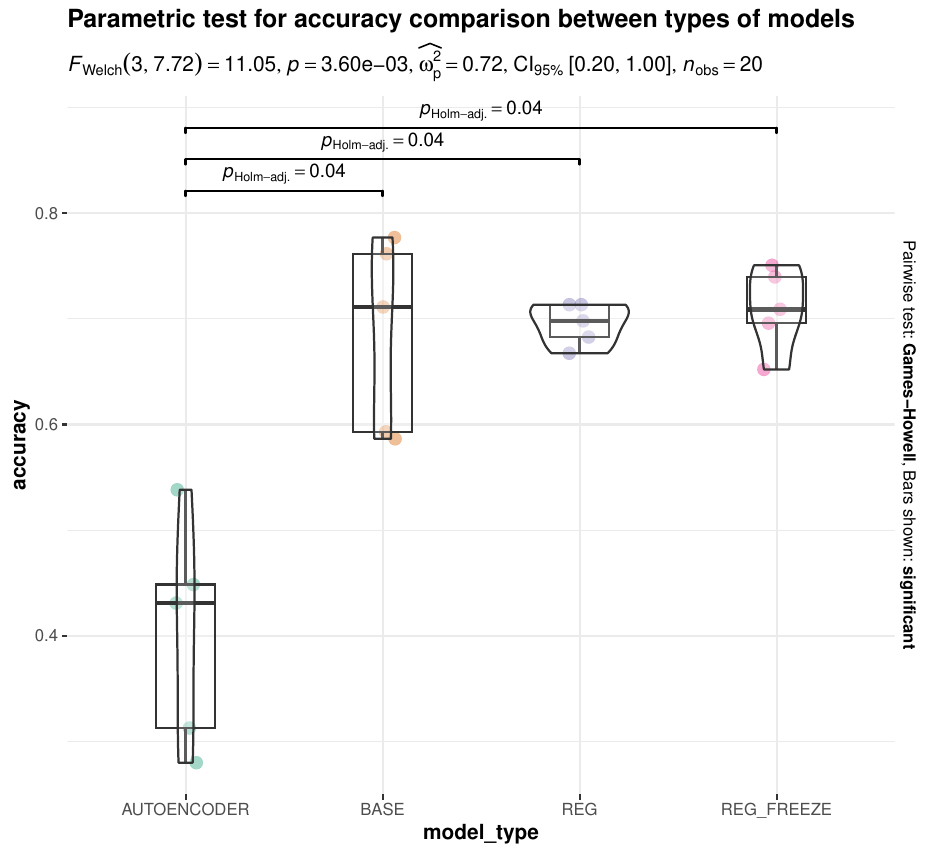}
        \label{fig:acc-anova}
    \end{subfigure}
    \begin{subfigure}{0.6\textwidth}
        \includegraphics[width=\textwidth]{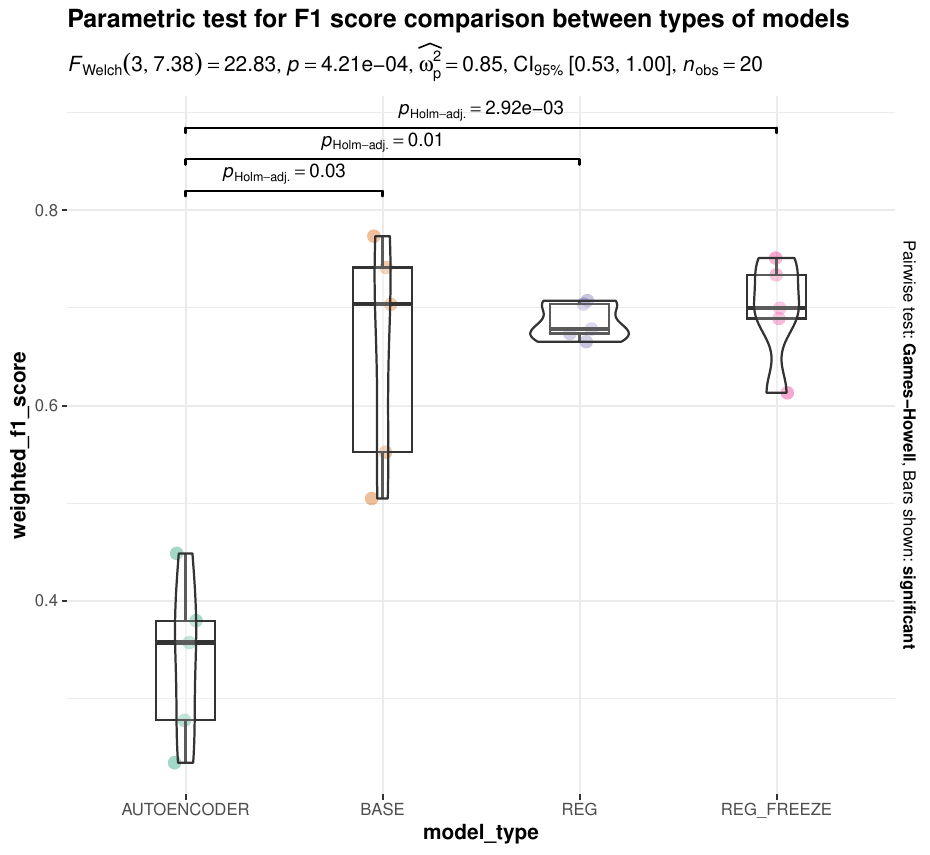}
        \label{fig:f1-anova}
    \end{subfigure}
    \caption{Results of the one-way ANOVA and the post-hoc tests for accuracy (top) and F1-score (bottom). A model is a composition of encoder-modeler and the best decision tree classifier. The normality of residuals was confirmed using the Shapiro-Wilk tests. Low $p$-values indicate that at least one type of models is significantly different in terms of performance. The Games-Howell post-hoc tests revealed significant differences between the Baseline (`AUTOENCODER') and all remaining architectures.}
    \label{figs: classification statistics}
\end{figure*}

Figure \ref{figs: classification statistics} summarizes the outcomes of statisical analysis with ANOVA and post-hoc Holm test, which confirm that our encoder-modeler-renderer models are better than the autoencoders on accuracy and weighed average of F1-score. 

\section{Attribute importance}

\begin{figure*}[t]  
\centering\includegraphics[width=0.7\linewidth, trim={0.2cm 0.2cm 0.2cm 0.8cm}, clip]{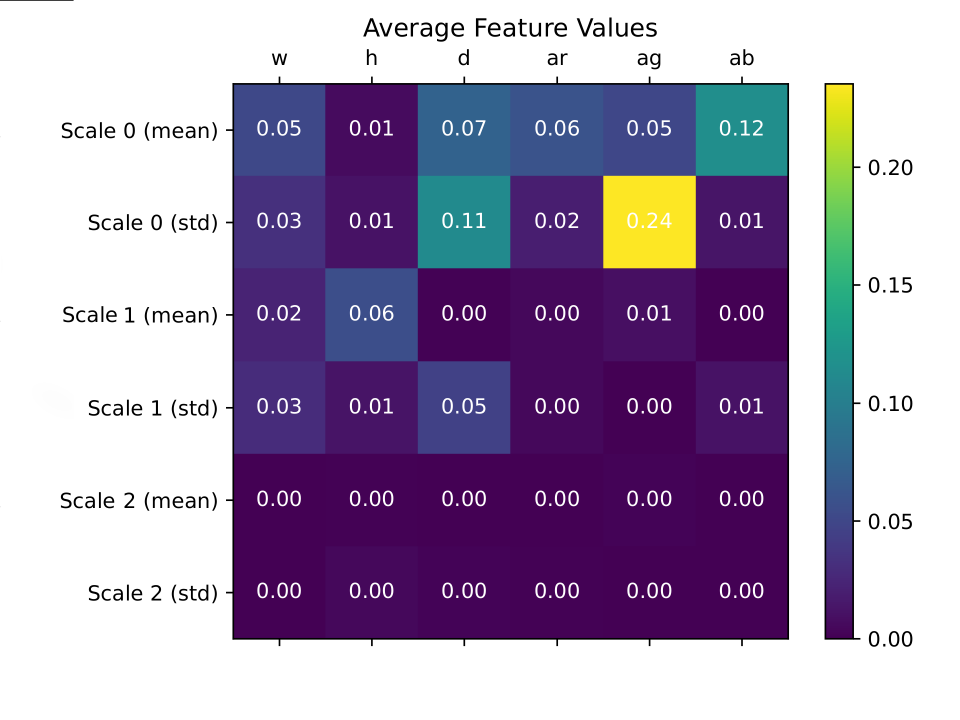}
    \caption{The importance of each of 36 attributes averaged over 15 decision trees obtained with \mname. The columns correspond to variables that control the rendering: the horizontal and vertical scaling ($w$ and $h$), rotation angle $d$, and RGB color components $a_r$, $a_g$, and $a_b$. The values of parameters from a particular scale are aggregated using mean and standard deviation (as seen in the rows). }
    \label{fig: features-importance_heatmap}
\end{figure*}

Figure \ref{fig: features-importance_heatmap} presents the heatmap of relative importance of attributes of visual primitives (in columns) extracted by \mname at particular spatial scales (in rows).  

\bibliographystyle{unsrtnat}
\bibliography{bib-bsc, bib-nesy24}  

\end{document}